\pdfoutput=1

\documentclass[11pt]{article}

\usepackage{emnlp2021}

\usepackage{times}
\usepackage{latexsym}

\usepackage[T1]{fontenc}

\usepackage[utf8]{inputenc}

\usepackage{microtype}

\usepackage{amsmath,amssymb,amsfonts}
\usepackage{booktabs}
\usepackage{graphicx}
\usepackage{subcaption}
\usepackage{algorithm} 
\usepackage{algpseudocode}
\usepackage{multirow}
\usepackage{makecell}
\usepackage{hyperref}
\usepackage{booktabs}

\title{Are Training Resources Insufficient? Predict First Then Explain!}

\author{\textbf{Myeongjun Jang} \and \textbf{Thomas Lukasiewicz} \\
    University of Oxford \\
    Department of Computer Science \\
    \texttt{\{myeongjun.jang, thomas.lukasiewicz\}@cs.ox.ac.uk}}

\date{}

\begin{document}
\maketitle
\begin{abstract}
Natural language free-text explanation generation is an efficient approach to train explain\-able language processing models for com\-mon\-sen\-se-knowledge-requiring tasks. The most predominant form of these models is the  \textit{explain-then-predict} (EtP) structure, which first generates explanations and uses them for making decisions. The performance of EtP models is highly dependent on that of the explainer by the nature of their structure. Therefore, large-sized explanation data are required to train a good explainer model. However, annotating explanations is expensive. Also, recent works reveal that free-text explanations might not convey sufficient information for decision making. These facts cast doubts on the effectiveness of EtP models. In this paper, we argue that the \textit{predict-then-explain} (PtE) architecture is a more efficient approach in terms of the modelling perspective. Our main contribution is twofold. First, we show that the  PtE structure is the most data-efficient approach when explanation data are lacking. Second, we reveal that the PtE structure is always more training-efficient than the EtP structure. We also provide experimental results that confirm the theoretical advantages.
\end{abstract}

\section{Introduction}

The accurate but black-box decision-making nature of deep neural networks has become a driving force to develop eXplainable Artificial Intelligence (XAI) models. In natural language processing (NLP), extracting \textit{rationales}, which is a short but sufficient part of an input to make a decision on its own, has been widely studied to provide interpretability \cite{lei2016rationalizing, bastings2019interpretable, yu2019rethinking, jain2020learning, paranjape2020information}. 

Although rationale extraction is a very useful method, there is a limitation to employ this approach in tasks that should link commonsense knowledge information to decisions, such as Commonsense QA (CQA) \cite{commonsenseQA} and natural language inference (NLI) \cite{SNLI}. Therefore, rather than extracting words with high importance, it is more desirable to provide free-text explanations to fill the knowledge gap \cite{ESNLI, CosE, wiegreffe2020measuring}.

The advent of free-text explanation datasets on commonsense knowledge tasks, such as e-SNLI \cite{ESNLI} and CosE \cite{CosE}, have paved the way for training models that can provide predictions with explanations. A~more natural way for training models is to explain the 
underlying reasoning first and to use the explanations to predict a label \cite{ESNLI}. Based on this property, many works proposed a model with an \textit{explain-then-predict} (EtP) structure \cite{ESNLI, latcinnik2020explaining, Nile, LiREx}.

However, garnering enough \textit{(input, label, explanation)} triples is essential to train EtP models. The problem is that annotating explanations is a much more demanding work than annotating labels. This property precludes the development of free-text explainable models for other languages where a large-sized explanation dataset is unavailable. 

Also, recent works cast doubt on the necessity of explanations for decision making:

\begin{enumerate}
  \item Current state-of-the-art models for CQA \cite{xu2020fusing} and NLI \cite{pilault2020conditionally} leverage no explanation information.
  \item In the work of \citet{Nile}, a \textit{post-hoc} model that predicts first and matches a corresponding explanation recorded a better NLI performance than EtP models.
  \item Through experiments, it has been shown that explanations might not convey sufficient information for decision making \cite{LiREx, wiegreffe2020measuring}.
  \item Joint models like WT5?! \cite{WT5}, which generate a label and explanation simultaneously, do not take an explanation as input but produce excellent results.
  \item \citet{wiegreffe2020measuring} showed that joint models generate more label-informed explanations compared to EtP models.
\end{enumerate}

Provided the effect of leveraging explanations for decision making is negligible, it would be beneficial to think beyond the EtP structure. 

In this paper, we argue that the \textit{predict-then-explain} (PtE) structure is a more efficient approach than joint models and the EtP structure in terms of the modelling perspective. Specifically, the PtE structure is data-efficient, i.e., it produces better results when explanation data are scarce, compared to EtP and joint models. Also, the PtE approach yields comparable but  better results as the EtP approach, while using less training time, i.e., it is more training-efficient. The main contributions of this paper can be briefly summarised as follows:

\begin{itemize}
    \item Through experiments, we reveal that PtE models are more data-efficient than joint and EtP models.
    \item We also show that PtE models are more trai\-ning-efficient compared to the EtP approach.
    \item We confirmed that the PtE approach has additional useful advantages: it generates more label-informed explanations and is able to produce label-conditioned explanations.
\end{itemize}

\section{Background}
In this section, we formulate the training objective of three structures: Joint, EtP, and PtE structures, and provide theoretical evidence for the effectiveness of the PtE structure.
\subsection{Problem Formulation}
Let the training data consist of $n$ triples $X=\{(\textrm{x}_1,\textrm{y}_1, \textrm{e}_1),\ldots, (\textrm{x}_n,\textrm{y}_n, \textrm{e}_n)\}$, where $\textrm{x}_i$ is a text input, $\textrm{y}_i$ is a target label, and $\textrm{e}_i$ is a corresponding explanation. The training objective of free-text explanation models is to maximise the conditional likelihood of the target labels and explanations given the input text:
\begin{equation}\label{equation.1}
\begin{gathered}
\mathcal{L}=\prod_{i=1}^{n} p(\textrm{y}_i, \textrm{e}_i|\textrm{x}_i).
\end{gathered}
\end{equation}

A joint model like WT5?! \cite{WT5}, which is designed to generate labels and explanations simultaneously, is trained to directly fit the above conditional likelihood.

By using the multiplication law of conditional probability, we can decompose Eq.~(\ref{equation.1}) into two different forms \cite{pruthi2020weakly}. The first form is as follows:
\begin{equation}\label{equation.2}
\begin{gathered}
\mathcal{L}=\prod_{i=1}^{n} \overbrace{(p(\textrm{e}_i| \textrm{x}_i)}^\text{explain} \overbrace{p(\textrm{y}_i|\textrm{x}_i, \textrm{e}_i)}^{\text{predict}}. 
\end{gathered}
\end{equation}

The likelihood function of Eq.~(\ref{equation.2}) corresponds to the objective of the EtP structure, which generates an explanation first, and leverages it for prediction. Alternatively, we can also decompose Eq.~(\ref{equation.1}) as follows:
\begin{equation}\label{equation.3}
\begin{gathered}
\mathcal{L}=\prod_{i=1}^{n} \overbrace{(p(\textrm{y}_i| \textrm{x}_i)}^\text{predict} \overbrace{p(\textrm{e}_i|\textrm{x}_i, \textrm{y}_i)}^{\text{explain}} .
\end{gathered}
\end{equation}

The above function corresponds to the PtE structure that first predicts a target label, and leverages it for generating an explanation.

\subsection{Data Efficiency}\label{sec.data_efficiency}
Now, let us consider that we have additional $m$ data points ($m \gg n$) that have only input text and target labels, i.e., $X^{\prime}=\{(\textrm{x}_1^{\prime}, \textrm{y}_1^{\prime}), \ldots, (\textrm{x}_m^{\prime}, \textrm{y}_m^{\prime})\}$. Note that this scenario can frequently occur in real-world applications due to the high cost and difficulty of data annotation \cite{cer2018universal, yang2019predicting}.

When it comes to joint models, only $n$ (out of $m+n$) data points are available for training, since explanations are essential to compute the likelihood in Eq.~(\ref{equation.1}). For a similar reason, the EtP structure also can only use $n$ data points, since both components of the likelihood function in Eq.~(\ref{equation.2}) require explanations \cite{pruthi2020weakly}.

On the contrary, the PtE structure requires explanations only for computing the \textit{explainer} component of the likelihood function. Thus, it can take advantage of weakly supervised learning, while using $m+n$ data points for training the \textit{predictor} and leveraging $n$ examples for training the \textit{explainer} \cite{pruthi2020weakly}. Therefore,  PtE models are more data-efficient than EtP and joint models.

\subsection{Training Efficiency}
The required training time for both PtE and EtP models is almost similar. They have nearly the same components of the likelihood function, i.e., an \textit{explainer} and a \textit{predictor}, which are marginally different in their inputs. However,  PtE models likely have a better performance than the EtP approach.  

A leading cause of this phenomenon is that the EtP structure takes explanations as inputs to the \textit{predictor}. Generating correct sentences is a much more difficult task than predicting correct labels. This causes a discrepancy between ground-truth and generated explanations. In addition, neural language generation models suffer from the \textit{exposure bias} problem, which refers to the train-test discrepancy that arises when the auto-regressive text generative model uses ground-truth contexts at the training stage but leverages generated ones at the inference stage \cite{schmidt2019generalization}. Although the language generation performance gap caused by the exposure bias might not be severe \cite{he2020quantifying}, the discrepancy between train-inference explanations should not be overlooked. This is because it might violate the $i.i.d$ condition, a fundamental assumption in machine learning, and hence degrade the performance of the \textit{predictor}. Therefore, the PtE structure is more training-efficient than the EtP approach, because it can produce a higher performance while using a similar training time.

\subsection{Semi-labeled EtP Structure}
A possible approach to deal with the EtP structure's issues, i.e., the exposure bias and the lack of explanation data, is to take advantage of semi-supervised learning. We can first train the \textit{explainer}, generate explanations for the whole data points, and use them during the \textit{predictor} training stage \cite{Nile, jain2020learning,  LiREx}. In this way, we can minimise the training-inference explanation discrepancy and use all labeled data points. The only drawback of this approach is the additionally required time to generate semi-labeled explanations. Therefore, provided the PtE structure produces comparable results with the semi-supervised EtP structure, we can say that the PtE approach is still more training-efficient than the EtP approach. We show that this statement holds in our experiments.

\begin{figure*}[t!]
	\centering
	\begin{subfigure}[b]{1.0\textwidth}
		\includegraphics[width=\linewidth]{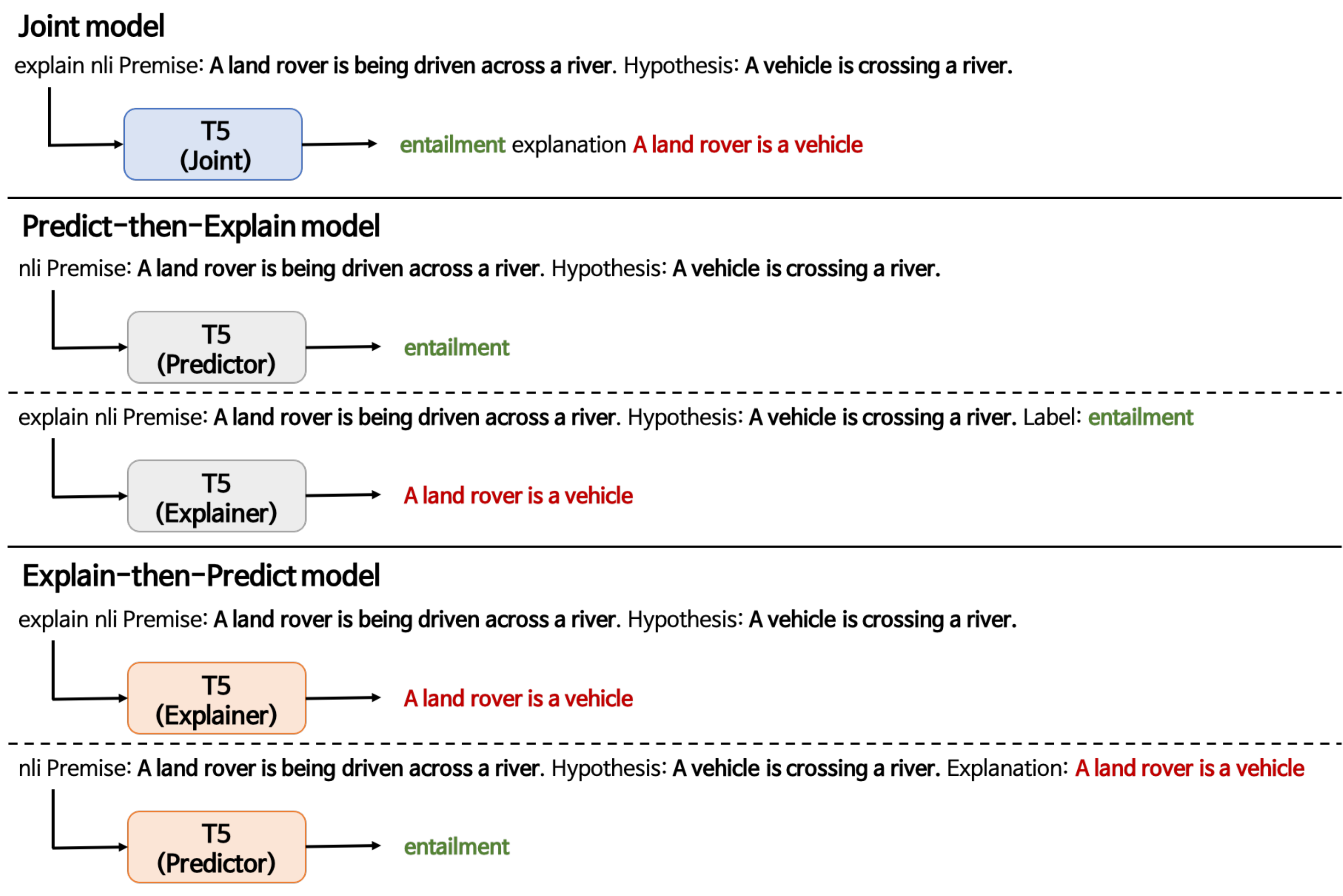}
	\end{subfigure} 
    \caption{Examples of the joint, the PtE, and the EtP architecture for   e-SNLI with a human-annotated rationale.}
    \label{fig:archi_example}
\end{figure*}

\section{Experiments}

\subsection{Dataset}
For the experiments, we used two datasets for commonsense-knowledge-requiring tasks: e-SNLI \cite{ESNLI} for the NLI task and CosE \cite{CosE} for the CQA task. Both datasets provide human-annotated free-text explanations. CosE has two versions: v1.0 and v1.11, which are different in the number of answer choices. The main task of e-SNLI is NLI, which aims to predict whether a hypothesis sentence entails, contradicts, or is neutral towards a premise sentence. A~single explanation is annotated to this task in training data. Development and test data contain 3 annotated explanations for each instance. The CosE dataset is designed to provide human annotated explanations for CQA task. This dataset has two versions: v1.0 and v1.11. The task is to predict the answer for a given question from 3 (v1.0) or 5 (v1.11) choices. The detailed statistics of each dataset are illustrated in Table \ref{table.dataset} in Appendix.

\subsection{Experimental Design}
\paragraph{Backbone model.} For a fair comparison, it is important to use the same architecture for the joint, the EtP, and the PtE model. Therefore, we used a Transformer-based text-to-text \cite{T5} approach, because this structure could be applied to all three models. We used the  T5-base model \cite{T5}, because it produced promising results on explanation generation \cite{WT5}. However, any kind of pre-trained generation model, such as GPT2 \cite{GPT2} and XLNet \cite{XLNet}, can be used.

\paragraph{Number of explanations.} To evaluate the data efficiency of the three structures, we designed an experimental environment illustrated in Section \ref{sec.data_efficiency}  using different amounts of explanation data. For the e-SNLI dataset, we used 10\%, 30\%, and 100\% of the total explanation data. As for CosE, we omitted the 10\% case, because the  remaining data seem not sufficient to train a large-sized model.

\paragraph{Implementation and training.} Figure \ref{fig:archi_example} briefly  illustrates the architectures for the joint, the PtE, and the EtP model. More detailed information regarding the input and the output format for each structure is presented in Table \ref{table.in_out_format} in Appendix \ref{sec:appendix}. For the implementation, we used the  transformers package\footnote{\href{https://github.com/huggingface/transformers}{https://github.com/huggingface/transformers}} provided by Hugging Face.

\medskip
It is worth to mention that our purpose is not to reproduce the state-of-the-art performance, but to compare results under the same conditions. Therefore, all three architectures are trained under the same training options. All models were trained for 20 epochs with a batch size of 8.\footnote{We used a single GeForce GTX TITAN XP GPU for training the models.} The training was ended if the validation-set loss had not decreased for 10 consecutive logging steps (10,000 iterations for e-SNLI and 100 iterations for CosE).  We used the AdamW \cite{AdamW} optimiser with learning rate $1e^{-4}$ and epsilon $1e^{-8}$. We also used gradient clipping to a maximum norm of 1.0 and a linear learning rate schedule decaying from $5e^{-5}$.

\paragraph{Evaluation metrics.} To evaluate the classification task performance, i.e., NLI and CQA, we used the accuracy. For measuring the quality of explanations, we used n-gram based metrics: BLEU \cite{BLEU}, METEOR \cite{METEOR}, ROUGE-L \cite{ROUGE},  and BERTScore \cite{BERTScore}, which is designed to measure the semantic similarity of two sentences. When it comes to the BLEU score, we used SacreBLEU v1.5.0 \cite{post2018call} with $\exp$ smoothing. Originally, the e-SNLI dataset contains 3 ground-truth explanations for the dev/test set. We used the first two gold explanations for the evaluation \cite{ESNLI, WT5}.

\section{Results}

\begin{table*}[t!]
	\begin{center}

		\caption{Results for the experiments on the e-SNLI dataset. The value written on the far left column indicates the ratio of used explanations. We trained each model 3 times and recorded the average of each metric. The best evaluation metric values are in bold. ``SL'' refers to using semi-labeled explanations for the EtP model.} \label{table.esnli_result}%
		\renewcommand{\arraystretch}{1.3}
		\footnotesize{
			\centering{\setlength\tabcolsep{4pt}
		\begin{tabular}{cl|ccccc|ccccc}
		\toprule
		\hline
		\multicolumn{2}{c|}{\multirow{2}{*}{\textbf{Model}}} & \multicolumn{5}{c|}{\textbf{Dev set}} & \multicolumn{5}{c}{\textbf{Test set}}\\
		& & Acc & BLEU & METEOR & Rouge-L & BERTScore 
		& Acc & BLEU & METEOR & Rouge-L & BERTScore \\\hline
		
        \multirow{4}{*}{\makecell{10\%}} 
        & Joint & 87.76 & 26.22 & \textbf{47.57} & 37.78 & 91.06 & 87.65 & 25.91 & 24.50 & 37.74 & 91.05 \\ 
        & EtP & 84.26 & 27.03 & 42.54 & 38.56 & \textbf{91.28} & 83.87 & 27.13 & 25.27 & \textbf{38.70} & \textbf{91.28}  \\ 
        & EtP (SL) & 90.51 & 27.03 & 42.54 & 38.56 & \textbf{91.28} & 90.16 & 27.13 & 25.27 & \textbf{38.70} & \textbf{91.28}  \\ 
        & PtE & \textbf{91.16} & \textbf{28.29} & 38.20 & \textbf{38.59} & \textbf{91.28} & \textbf{90.81} & \textbf{27.83} & \textbf{40.11} & 38.66 & 91.26 \\  \hline

        \multirow{4}{*}{\makecell{30\%}} 
        & Joint & 89.61 & 27.35 & 36.11 & 38.58 & 91.26 & 89.14 & 27.05 & 20.01 & 38.50 & 91.19 \\ 
        & EtP & 87.48 & 28.56 & 36.04 & \textbf{39.22} & \textbf{91.41} & 87.12 & 28.24 & 23.16 & 39.29 & 91.40 \\ 
        & EtP (SL) & 90.78 & 28.56 & 36.04 & \textbf{39.22} & \textbf{91.41} & 90.21 & 28.24 &  23.16 & 39.29 & 91.40 \\ 
        & PtE & \textbf{91.16} & \textbf{29.11} & \textbf{41.88} & 39.15 & \textbf{91.41} & \textbf{90.81} & \textbf{28.91} & \textbf{36.16} & \textbf{39.35} & \textbf{91.46}\\ \hline

        \multirow{4}{*}{\makecell{100\%}} 
        & Joint & \textbf{91.47} & 28.35 & \textbf{33.83} & 39.44 & 91.41 & \textbf{90.91} & 27.68 & 23.04 & 39.26 & 91.37 \\ 
        & EtP & 90.01 & 29.86 & 32.71 & \textbf{40.11} & \textbf{91.61} & 89.22 & 29.07 & 25.04 & \textbf{39.91} & 91.56 \\ 
        & EtP (SL) & 90.95 & 29.86 & 32.71 & \textbf{40.11} & \textbf{91.61} & 90.34 & 29.07 & 25.04 & \textbf{39.91} & 91.56 \\ 
        & PtE & 91.16 & \textbf{29.87} & 33.61 & 39.97 & 91.60 & 90.81 & \textbf{29.36} & \textbf{36.65} & 39.82 & \textbf{91.59} \\ \hline

		\bottomrule
		\end{tabular}}}
	\end{center}
\end{table*}

\begin{table*}[t!]
	\begin{center}
		\caption{Results for the experiments on the CosE dataset. The value written on the far left column indicates the ratio of used explanations. We trained each model 3 times and recorded the average of each metric. The best evaluation metric values are in bold. ``SL'' refers to using semi-labeled explanations for the EtP model.} \label{table.cose_result}%
		\renewcommand{\arraystretch}{1.3}
		\footnotesize{
			\centering{\setlength\tabcolsep{4pt}
		\begin{tabular}{cl|ccccc|ccccc}
		\toprule
		\hline
		\multicolumn{2}{c|}{\multirow{2}{*}{\textbf{Model}}} & \multicolumn{5}{c|}{\textbf{v1.0 Dev set}} & \multicolumn{5}{c}{\textbf{v1.11 Dev set}}\\ 
		& & Acc & BLEU & METEOR & Rouge-L & BERTScore 
		& Acc & BLEU & METEOR & Rouge-L & BERTScore \\ \hline

        \multirow{4}{*}{\makecell{30\%}} 
        & Joint &  61.19 & 5.08 & 10.88 & 21.00 & 85.58 & 50.10 & 4.98 & 9.06 & 14.03 & 83.60 \\ 
        & EtP & 54.95 & 5.39 & 11.12 & 20.91 & 85.83 & 43.68 & \textbf{5.37} & 8.70 & 13.84 & 83.57  \\ 
        & EtP (SL) & 62.07 & 5.39 & 11.12 & 20.91 & 85.83 & 55.31 & \textbf{5.37} & 8.70 & 13.84 & 83.57 \\ 
        & PtE & \textbf{65.05} & \textbf{6.40} & \textbf{11.94} & \textbf{22.09} & \textbf{85.99} & \textbf{57.74} & 5.16 & \textbf{9.42} & \textbf{14.42} & \textbf{83.71} \\ \hline
        
        \multirow{4}{*}{\makecell{100\%}} 
        & Joint & 64.11 & 5.35 & 11.12 & 21.38 & 85.79 & 53.43 & 4.96 & 9.46 & 14.79 & \textbf{83.72} \\
        & EtP & 55.96 & 5.48 & 11.09 & 20.69 & 85.96 & 46.93 & 5.10 & 8.32 & 13.80 & 83.45  \\ 
        & EtP (SL) & \textbf{65.13} & 5.48 & 11.09 & 20.69 & 85.96 & 55.94 & 5.10 & 8.32 & 13.80 & 83.45 \\ 
        & PtE & 65.05 & \textbf{6.51} & \textbf{12.07} & \textbf{22.36} & \textbf{86.10} & \textbf{57.74} & \textbf{6.60} & \textbf{11.17} & \textbf{16.75} & 83.52 \\ \hline
        
		\bottomrule
		\end{tabular}}}
	\end{center}
\end{table*}

The experimental results for the e-SNLI and CosE datasets are summarised in Tables \ref{table.esnli_result} and  \ref{table.cose_result}, respectively.  For generating explanations, we greedy-decode for 100 tokens or until an EOS token appears. Note that the training data size of CosE v1.0 and v1.1 are much smaller than e-SNLI. They are less than 10,000 instances for each even when 100\% of explanation data are used.

\subsection{Results for Data Efficiency}
To support our claim  that the PtE structure is the most data-efficient method, we first compared the results of the joint, the EtP, and the PtE model. When explanation data are plenty, e.g., e-SNLI with 100\% explanations, the joint model recorded the best main task performance. However, for the cases where explanation data are insufficient, the PtE model performed the best in general. Also, we ascertained that the PtE model generates similar or better quality explanations compared to the other methods while recording the best figure in 30 out of 40 cases.

We observed that the accuracy of the joint and the EtP model degrade along with the decrease of the size of explanation data. Also, the decrease rate of the EtP structure is more rapid than of  the joint model. We conjecture that  the additional factor for degradation is the exposure bias that only the EtP structure suffers from. This can  be inferred from the results of the EtP and the EtP (SL) model when 100\% of explanation data are used. In that case, the only difference between the two models is whether explanations are ground-truth or generated ones. The result of EtP with the semi-labeled explanation model performs much better than the original EtP model, which implies that the exposure bias issue exists. On the contrary, the accuracy of the PtE model is not affected by the number of explanations due to its structural nature. This could be a useful property for training the models for other languages, such as German, French, and Korean, where large-sized explanation data are not yet available.
 
\begin{table*}[t!]
	\begin{center}
		\caption{Results for the label-informedness experiments. We recorded the average accuracy of 3 repetitions. The best values among generated explanations are in bold. The value written in parentheses denotes the performance recover ratio. Note that the  generated explanations for the EtP and the EtP (SL) model are the same.} \label{table.label_inform}%
		\renewcommand{\arraystretch}{1.3}
		\footnotesize{
			\centering{\setlength\tabcolsep{7pt}
		\begin{tabular}{c|cc|cc}
		\toprule
		\hline
		\multirow{2}{*}{\textbf{Model}} & \multicolumn{2}{c|}{\textbf{e-SNLI}} & \multicolumn{2}{c}{\textbf{CosE}} \\ 
		& Dev & Test
		& v1.0 & v1.11 \\ \hline
		
		$R^{*} \rightarrow L$ & 97.81 & 97.90 & 86.63 & 69.12 \\
		$R_{joint} \rightarrow L$ & 90.01 (92.03\%) & 89.74 (91.66\%) & 60.91 (70.31\%) & 52.01 (75.24\%)  \\
		$R_{EtP} \rightarrow L$ & 89.08 (91.08\%) & 88.27 (90.16\%) & 51.61 (59.58\%) & 36.15 (52.29\%)  \\
		$R_{PtE} \rightarrow L$ & \textbf{90.31} (92.33\%) & \textbf{89.94} (91.87\%) & \textbf{61.54} (71.04\%) & \textbf{57.99} (83.89\%)  \\ \hline
		
		\bottomrule
		\end{tabular}}}
	\end{center}
\end{table*}

\subsection{Results for Training Efficiency}
Our second claim is that the PtE model is always more training-efficient than the EtP model. We confirmed that this statement holds by comparing the performance of PtE, EtP, and EtP with semi-labeled explanations.

First, we found that the PtE model performs the main task significantly better than the EtP structure, even though both models consume almost identical training time. As illustrated before, the leading causes of this outcome are the EtP structure's dependence on the explanation data and the  exposure bias problem.

We observed that leveraging semi-labeled explanations for the EtP model is of benefit to improve the performance. The accuracy increased by 12\% on average, and the model shows stable results regardless of the number of leveraged explanations. Also, the semi-labeled EtP model even recorded the highest accuracy on CosE v1.0  when 100\% of explanations are used, but the difference with the PtE model is marginal.

Overall, the PtE model produced a  comparable but better performance than the EtP model with semi-labeled explanations, even though the latter requires additional time for generating semi-labeled explanations. As more data points with label but no explanations are available, the required time for training increases considerably.\footnote{For instance, it took more than 4 hours to generate explanations for e-SNLI training data on our computing resource.} Therefore, we can say that the PtE structure is always more  training-efficient than the EtP model.

\subsection{Label Informedness}
Although the PtE method produced better results than the others in terms of the quality of explanations in many cases, it is hard to conclude that it generates the best-quality explanations due to the limitation of automatic evaluation metrics \cite{ESNLI}. Therefore, we additionally conducted an experiment proposed by \citet{wiegreffe2020measuring} for a more detailed evaluation.

The experiment is designed to measure label-explanation association, a necessary property of faithful explanations \cite{wiegreffe2020measuring}. The overall procedure of the experiment is described below. For convenience, we denote $R \rightarrow L$ as a model that predicts the target label by only leveraging the explanation.

\begin{enumerate}
  \item Train a $R \rightarrow L$ model by leveraging the gold-standard explanation ($R^*$);
  \item Evaluate the accuracy of $R^* \rightarrow L$;
  \item Train $R \rightarrow L$ models by leveraging the generated explanations of each model ($\hat{R}$);
  \item Evaluate the accuracy of $\hat{R} \rightarrow L$ and measure the ability to recover $R^* \rightarrow L$ model's performance.
\end{enumerate}

The results are summarised in Table \ref{table.label_inform}. It is observed that the PtE model recovers the ground-truth ($R^{*}$) performance the best. The joint model showed almost comparable results but marginally below the PtE model's performance. On the contrary, the explanations from the EtP model fall short compared to other two models. This suggests that adding the label signal is of benefit to generate label-informed explanations, which can avoid cascading errors caused by less informative free-text explanations \cite{wiegreffe2020measuring}.

\begin{table*}[t!]
	\begin{center}
		\caption{Examples of label-conditioned explanations of the PtE model on wrongly-classified e-SNLI instances. The explanations written in the left column are  generated ones conditioned on the true label, while those written in the right column are  generated explanations for the predicted label.} \label{table.cond_examples}%
		\renewcommand{\arraystretch}{1.7}
		\footnotesize{
			\centering{\setlength\tabcolsep{5pt}
		\begin{tabular}{p{7.5cm}|p{7.5cm}}
		\toprule
		\hline
		
		\multicolumn{2}{l}{\textsc{Premise}: A person with a purple shirt is painting an image of a woman on a white wall.} \\
		\multicolumn{2}{l}{\textsc{Hypothesis}: A woman paints a portrait of a person.} \\ 
		\textsc{True Label}: Neutral & \textsc{Predicted Label}: Entailment  \\
		\textsc{Explanation}: Not all images are a portrait. & \textsc{Explanation}: A portrait is an image.  \\ \hline

		\multicolumn{2}{l}{\textsc{Premise}: A bird is flapping its wings on the water.} \\
		\multicolumn{2}{l}{\textsc{Hypothesis}: The duck is swimming in the lake.} \\ 		
		\textsc{True Label}: Neutral & \textsc{Predicted Label}: Contradiction  \\
		\textsc{Explanation}: Not all birds are a duck. Flapping its wings does not imply swimming. & \textsc{Explanation}: A bird is not a duck. \\ \midrule
		
		\multicolumn{2}{l}{\makecell[l]{\textsc{Premise}: One tan girl with a wool hat is running and leaning over an object, while another person in a wool hat is sitting \\ on the ground.}} \\
		\multicolumn{2}{l}{\textsc{Hypothesis}: A man watches his daughter leap.} \\ 		
		\textsc{True Label}: Neutral & \textsc{Predicted Label}: Contradiction  \\
		\textsc{Explanation}: The girl is not necessarily the man's daughter. & \textsc{Explanation}: A girl is not a man. \\ \midrule
		\bottomrule
		\end{tabular}}}
	\end{center}
\end{table*}

\subsection{Label-Conditioned Explanations}
Another useful structural benefit of the PtE model is that it allows to generate explanations for a specific class by conditioning the \textit{explainer} on the task label. In Table \ref{table.cond_examples}, we attached three examples of label-conditioned explanations of the PtE model on wrongly-classified e-SNLI instances. More examples for the e-SNLI and the CosE dataset are available in Tables  \ref{table.more_cond_examples}, \ref{table.more_cond_examples_cos1.0}, and \ref{table.more_cond_examples_cos1.11} in Appendix \ref{sec:appendix}.

This characteristic is especially useful when the model is less confident on its prediction \cite{pruthi2020weakly}. Also, exploring the outcome for each class could help us understand the model's behaviour and draw meaningful insights for improving the performance. For instance, through examples in Table \ref{table.cond_examples}, we can infer that the model's \textit{predictor} has more issues than the \textit{explainer},  because explanations for true labels seem reasonable.

Unlike the PtE approach, the joint and the EtP model can not generate label-conditioned explanations in principle. Although the EtP model can provide label-conditioned explanations with some modification, it requires additional components, or it is only applicable to a certain task. For instance, NILE \cite{Nile} trains distinct explanation generators for each class of the e-SNLI dataset, i.e., entailment, contradiction, and neutral. Although this model can provide label-conditioned explanations, it could not be applied to the CosE dataset due to the numerous answer choices of the CQA task, not to mention the increased time for training independent explanation generators. LIREX \cite{LiREx} can also provide explanations for all labels, but it has to first train a label-aware rationale extractor, which requires a token-level annotated dataset. 

\section{Related Work}
An increasing amount of work has been proposed to provide explanations for the model's behaviour. In NLP, the most widely used approach is \textit{rationale extraction}, which extracts important words or tokens for making a decision \cite{lei2016rationalizing, bastings2019interpretable, yu2019rethinking, jain2020learning, paranjape2020information}. These models generally have two components: an \textit{extractor} that extracts rationales from an input sentence, and a \textit{predictor} that conducts a  prediction by only using extracted rationales.  An alternative method is providing natural language, free-text explanations, which is a more comprehensive and user-friendly form of explainability \cite{camburu2020make}.

\paragraph{Explain-then-Predict.} It is a natural approach for predicting an answer to think of the explanation first and make a decision based on the explanation. For this reason, EtP structures have been the most widely used framework for providing free-text explanations. \citet{ESNLI} first applied this structure to the NLI task. Similarly, \citet{CosE} implemented an EtP model for the CQA task, which takes questions and answer choices as an input for generating explanations. \citet{Nile} proposed a framework named NILE for the NLI task. This model first generates explanations for each label. These explanations are fed into the explanation processor to select the most reasonable explanation. \citet{LiREx} proposed a framework called LiREx, which is similar to  NILE but has an additional component. This model first extracts a label-aware token-level rationale and employs them for generating explanations for each label. The explanations are then leveraged for prediction. Unlike the aforementioned framework-based approaches that proceed step-by-step, \citet{latcinnik2020explaining} proposed a fully differentiable EtP model. To overcome the non-differentiable loss due to the discrete symbols of the explanation generator, \citet{latcinnik2020explaining} used a Gumbel-Softmax estimator \cite{jang2016categorical}.

\paragraph{Jointly Predict-and-Explain.} \citet{T5} proposed a unified transfer learning technique, called T5, which transforms every natural language problem into a text-to-text format. This approach has enabled the training of a jointly predicting and explaining model in a considerably easy way. The WT5 model \cite{WT5} leveraged T5 for commonsense requiring explanation generation tasks. This model showed promising results, and especially the model based on the T5-11B model recorded a state-of-the-art performance. \citet{wiegreffe2020measuring} found that joint predict-and-explain models generate more label-informed and faithful explanations compared to the EtP framework. \citet{li2021you} proposed another form of joint model. Unlike WT5, they separated the \textit{predictor} and the  \textit{explainer}. To jointly train both components, they introduced two additional objective functions: the guide loss that forces the \textit{explainer} to generate consistent outputs that coordinate with the \textit{predictor}, and the discrepancy loss that connects the original loss for the \textit{predictor} and the guide loss.

\paragraph{Predict-then-Explain.} This approach generates label-conditioned explanations, and hence is considered as a \textit{post-hoc} algorithm. We can train a PtE model in two ways. First, we can minimise the classification loss and the generation loss simultaneously \cite{ESNLI}. Alternatively, training separated ``predictor'' and ``explainer'' is also possible \cite{CosE}. In this paper, we show that the PtE structure has more advantages than the EtP approach in terms of the modelling perspective, despite the mechanism of the EtP model sounding more natural. Similarly to our work, \citet{pruthi2020weakly} ascertained that, when it comes to  rationale extraction, the \textit{classify-then-extract} method has more benefits than the \textit{extract-then-classify}, which is a predominant structure for rationale extraction.

\section{Summary and Outlook}
In this paper, we show that the PtE model is the most efficient method in terms of the modelling perspective. Unlike the joint and the EtP model, the PtE model performs better even when explanation data are insufficient, hence it is more data-efficient. Also, we showed that the PtE model is always more training-efficient compared to the EtP model. The PtE model is free from not only the exposure bias but also generating semi-labeled explanations. Furthermore, the PtE model is able to generate both label-informed and label-conditioned explanations, which is a very useful characteristic.

Our work presents an opposite perspective to an often-cited claim that leveraging explanations for the model's decision-making is a more natural way, especially when explanation data are insufficient. In fact, the statement may not be true considering the human decision-making process. \citet{soon2008unconscious} suggest that our actions are initiated unconsciously long before we become aware that we have made a decision, and we are not even aware of our own mental process. They claim that an experience of consideration before making a decision is merely an illusion. Even if the statement is true, our experimental results suggest that current neural language models hardly exploit the benefits of explanations for decision making. Unlike these models, humans can provide reasonable explanations regardless of the size of training data, because we mostly make a decision based on our background knowledge. This suggests that there still exists a huge gap between neural language models and human language recognition. Hence, a thorough analysis of the decision-making process and algorithms is needed to develop the field of~XAI.

As future work, we aim to develop a method to remove spurious correlations between labels and explanations, because introducing the label signal can increase the risk of generating forged explanations \cite{LiREx}. Also, using commonsense knowledge graphs for reasoning and generating explanations is another exciting research direction.

\clearpage
\bibliography{emnlp2021}
\bibliographystyle{acl_natbib}
\clearpage

\appendix

\section{Appendix}
\label{sec:appendix}
\subsection{T5 Structure}
The T5 model \cite{T5} is a pre-trained Transformer model that unified every natural language task to text-to-text format.  This model is pretrained on multi-task language tasks, including machine translation, text classification, and question answering. Originally, the model is trained for all tasks at once by using a task-specific input-output format and task conditioning. However, we trained a separate model for each dataset and each case for the percentage of used explanations to simulate the situation where explanation data are insufficient.

\clearpage
\begin{table*}[t!]
	\begin{center}
		\caption{Statistics on datasets used for the experiments. For the sentence length information, mean and standard deviation of train/dev/test sets are provided. For an input sentence, we used the concatenation of premise and hypothesis for e-SNLI and a question for cosE. Note that CosE v1.0 and v1.11 do not contain a test set.} \label{table.dataset}%
		\renewcommand{\arraystretch}{1.7}
		\footnotesize{
			\centering{\setlength\tabcolsep{5pt}
				\begin{tabular}{cccc}
					\toprule
					\hline
					\textbf{Dataset} & \textbf{Num.Data points} & \textbf{Input Length} & \textbf{Free-text Explanation Length}\\
					& Train-Dev-Test & Num. Tokens & Num. Tokens \\
					\hline
					e-SNLI & 549,367-9,842-9,824 & 26.48$\pm$9.21 & 15.77 $\pm$ 8.08 \\
					CosE v1.0 & 7,610-950-None & 17.17 $\pm$ 7.40 & 16.15$\pm$10.38 \\
					CosE v1.11 & 9,741-1,221-None & 16.79$\pm$7.18 & 9.00$\pm$5.32 \\
					\hline
					\bottomrule	
		\end{tabular}}}
	\end{center}
\end{table*}

\begin{table*}[t!]
	\begin{center}

		\caption{Input/output formatting for joint, PtE, and EtP structures.} \label{table.in_out_format}
		\renewcommand{\arraystretch}{2.0}
		\footnotesize{
			\centering{\setlength\tabcolsep{5pt}
		\begin{tabular}{ll|l}
		\toprule
		\hline
		\multicolumn{3}{l}{\textbf{The joint model's input/output format}} \\ \hline
		\multicolumn{2}{c|}{\multirow{2}{*}{\makecell{Input}}}
		& CosE: explain cos Question: [$question$] Choice: [$choice_0$] Choice: [$choice_1$] Choice: [$choice_2$] \\
		& & e-SNLI: explain nli Premise: [$premise$] Hypothesis: [$hypothesis$] \\ 
	    \multicolumn{2}{c|}{Output} & [$label$] explanation [$expl$] \\ \hline
	    
	    \multicolumn{3}{l}{\textbf{The Predict-then-Explain model's input/output format}} \\
	    \multicolumn{3}{l}{\textbf{Predictor:}} \\ \hline
	    \multicolumn{2}{c|}{\multirow{2}{*}{\makecell{Input}}}
		& CosE: cos Question: [$question$] Choice: [$choice_0$] Choice: [$choice_1$] Choice: [$choice_2$] \\
		& & e-SNLI: nli Premise: [$Premise$] Hypothesis: [$hypothesis$] \\ 
	    \multicolumn{2}{c|}{Output} & [$label$] \\ \hline
	    \multicolumn{3}{l}{\textbf{Explainer:}} \\ \hline
	    \multicolumn{2}{c|}{\multirow{2}{*}{\makecell{Input}}}
		& CosE: explain cos Question: [$question$] Choice: [$choice_0$] Choice: [$choice_1$] Choice: [$choice_2$] Label: [$label$] \\
		& & e-SNLI: explain nli Premise: [$Premise$] Hypothesis: [$hypothesis$] Label: [$label$] \\ 
	    \multicolumn{2}{c|}{{Output}} & [$expl$] \\ 
	    \hline
	    
	    \multicolumn{3}{l}{\textbf{The Explain-then-Predict model's input/output format}} \\
	    \multicolumn{3}{l}{\textbf{Explainer:}} \\ \hline
	    \multicolumn{2}{c|}{\multirow{2}{*}{\makecell{Input}}}
		& CosE: explain cos Question: [$question$] Choice: [$choice_0$] Choice: [$choice_1$] Choice: [$choice_2$] \\
		& & e-SNLI: explain nli Premise: [$Premise$] Hypothesis: [$hypothesis$] \\ 
	    \multicolumn{2}{c|}{Output} & [$expl$] \\ \hline
	    \multicolumn{3}{l}{\textbf{Predictor:}} \\ \hline
	    \multicolumn{2}{c|}{\multirow{2}{*}{\makecell{Input}}}
		& CosE: cos Question: [$question$] Choice: [$choice_0$] Choice: [$choice_1$] Choice: [$choice_2$] Explanation: [$expl$] \\
		& & e-SNLI: nli Premise: [$Premise$] Hypothesis: [$hypothesis$] Explanation: [$explanation$] \\ 
	    \multicolumn{2}{c|}{{Output}} & [$label$] \\ 

	    \hline
		\bottomrule
		\end{tabular}}}
	\end{center}
\end{table*}

\begin{table*}[t!]
	\begin{center}
		\caption{More examples of label-conditioned explanations of PtE model on wrongly-classified e-SNLI instances. The explanations written in left column is a generated one conditioned on the true label while that written in right column is a generated explanation for predicted label.} \label{table.more_cond_examples}%
		\renewcommand{\arraystretch}{1.7}
		\footnotesize{
			\centering{\setlength\tabcolsep{5pt}
		\begin{tabular}{p{7.5cm}|p{7.5cm}}
		\toprule
		\hline
		
		\multicolumn{2}{l}{\textsc{Premise}: This child is on the library steps.} \\
		\multicolumn{2}{l}{\textsc{Hypothesis}: The child is on the steps inside the library.} \\ 
		\textsc{True Label}: Neutral & \textsc{Predicted Label}: Entailment  \\
		\textsc{Explanation}: The steps are not necessarily inside the library. & \textsc{Explanation}: The steps inside the library are library steps.  \\ \hline

		\multicolumn{2}{l}{\textsc{Premise}:Six dogs swimming in a river.} \\
		\multicolumn{2}{l}{\textsc{Hypothesis}: Six dogs swim across the river.} \\ 		
		\textsc{True Label}: Neutral & \textsc{Predicted Label}: entailment  \\
		\textsc{Explanation}: The dogs could be swimming in the river, not across the river. Flapping its wings does not imply swimming. & \textsc{Explanation}: The dogs are swimming across the river. \\ \midrule
		
		\multicolumn{2}{l}{\makecell[l]{\textsc{Premise}: A man is sitting on the floor, sleeping.}} \\
		\multicolumn{2}{l}{\textsc{Hypothesis}: A man is lying down, sleeping.} \\ 		
		\textsc{True Label}: Contradiction & \textsc{Predicted Label}: Entailment  \\
		\textsc{Explanation}: Sitting on the floor is a way of lying down. & \textsc{Explanation}: A girl is not a man. \\ \midrule
		\bottomrule
		\end{tabular}}}
	\end{center}
\end{table*}

\begin{table*}[t!]
	\begin{center}
		\caption{More examples of label-conditioned explanations of PtE model on wrongly-classified CosE v1.0 instances. The explanations written in left column is a generated one conditioned on the true label while that written in right column is a generated explanation for predicted label.} \label{table.more_cond_examples_cos1.0}%
		\renewcommand{\arraystretch}{1.7}
		\footnotesize{
			\centering{\setlength\tabcolsep{5pt}
		\begin{tabular}{p{7.5cm}|p{7.5cm}}
		\toprule
		\hline
		
		\multicolumn{2}{l}{\textsc{Question}: A fencing thrust with a sharp sword towards a person would result in what?} \\
		\textsc{True Label}: puncture wound & \textsc{Predicted Label}: injury  \\
		\textsc{Explanation}: a puncture wound is the only wound that would result from a sword thrust. & \textsc{Explanation}: a sword is a weapon that is used to protect against injury.  \\ \hline

		\multicolumn{2}{l}{\textsc{Question}: Brawn opened the curtains so that the sun could do what?} \\
		\textsc{True Label}: warm room & \textsc{Predicted Label}: warm house  \\
		\textsc{Explanation}: the sun is in the room. & \textsc{Explanation}: the sun is warm and the curtains are open to the sun. \\ \midrule
		
		\multicolumn{2}{l}{\textsc{Question}: How would you express information if you do not have a pen or pencil?} \\
		\textsc{True Label}: talk & \textsc{Predicted Label}: write down  \\
		\textsc{Explanation}: you can't write down or disagree, but you can talk. & \textsc{Explanation}: writing down is the only way to express information. \\ \midrule
		\bottomrule
		\end{tabular}}}
	\end{center}
\end{table*}

\begin{table*}[t!]
	\begin{center}
		\caption{More examples of label-conditioned explanations of PtE model on wrongly classified CosE v1.11 instances. The explanations written in left column is a generated one conditioned on the true label while that written in right column is a generated explanation for predicted label.} \label{table.more_cond_examples_cos1.11}%
		\renewcommand{\arraystretch}{1.7}
		\footnotesize{
			\centering{\setlength\tabcolsep{5pt}
		\begin{tabular}{p{7.5cm}|p{7.5cm}}
		\toprule
		\hline
		
		\multicolumn{2}{l}{\textsc{Question}: Applying for a job can make someone feel what sort of emotion, even if they get it?} \\
		\textsc{True Label}: stress & \textsc{Predicted Label}: frustration  \\
		\textsc{Explanation}: stress is a feeling. & \textsc{Explanation}: this word is most relevant.  \\ \hline

		\multicolumn{2}{l}{\textsc{Question}: He used an umbrella while tanning, where was he likely?} \\
		\textsc{True Label}: beach & \textsc{Predicted Label}: seattle  \\
		\textsc{Explanation}: beach is a place to relax. & \textsc{Explanation}: seattle is the only place where he is likely to use an umbrella. \\ \midrule
		
		\multicolumn{2}{l}{\textsc{Question}: The rats were hiding in the house, where were they?} \\
		\textsc{True Label}: cellar & \textsc{Predicted Label}: basement  \\
		\textsc{Explanation}: cellars are a place where rats are kept. & \textsc{Explanation}: basement is the only place where rats are hiding. \\ \midrule
		\bottomrule
		\end{tabular}}}
	\end{center}
\end{table*}

\end{document}